\documentclass[letterpaper, 10 pt, conference]{ieeeconf}
\IEEEoverridecommandlockouts
\usepackage[utf8]{inputenc}

\usepackage{array}
\usepackage{booktabs}
\usepackage{multirow}
\usepackage{siunitx}
\usepackage{graphicx}
\usepackage{amsmath}
\usepackage{amssymb}
\usepackage[T1]{fontenc}
\usepackage[font=footnotesize, labelfont=bf]{caption}

\DeclareSIUnit{\pixel}{px}
\DeclareSIUnit{\px}{px}

\usepackage{xcolor}
\definecolor{linkblue}{rgb}{0.10,0.30,0.65}
\definecolor{citegreen}{rgb}{0.10,0.45,0.25}
\definecolor{urlred}{rgb}{0.55,0.15,0.15}
\usepackage{hyperref}
\hypersetup{
  colorlinks=true,
  linkcolor=linkblue,    
  citecolor=citegreen,   
  urlcolor=urlred,
  bookmarksnumbered=true
}
\renewcommand{\baselinestretch}{0.97}

\title{\bf RadGeoLoc: Radar Geolocalization via Particle Filtering}

\author{Shaunak Kolhe$^{1}$, Peng Jiang$^{1}$, Kasi Viswanath$^{1}$, Srikanth Saripalli$^{1}$
\thanks{J. Mike Walker ’66 Department of Mechanical Engineering,
Texas A\&M University, College Station, TX 77840, USA %
        {\tt\small kolheshaunak@tamu.edu}, %
        {\tt\small maskjp@tamu.edu}, %
        {\tt\small kasiv@tamu.edu}, %
        {\tt\small ssaripalli@tamu.edu}} %
}

\begin{document}
\maketitle
\thispagestyle{empty}
\pagestyle{empty}

\begin{abstract}
 In real-time autonomous field applications, GNSS is often unreliable and inaccurate. A possible solution is localizing against a prebuilt map like a satellite image, as it is georeferenced and already surveyed. In this work, we localize a spinning radar measurement against overhead RGB imagery with a learned observation model. We use a radar representation pretrained by masked reconstruction and align it to features from a satellite image. A particle filter propagated by radar odometry proposes candidate poses, and our observation model provides a likelihood over these poses. We evaluate our method over 31 distinct routes, over 8 datasets spanning both urban and offroad conditions using only ground radar and a coarse local prior. We also examine the robustness over changing weather using repeated traversals over the same route over 13 months, in both urban and offroad conditions. 
 
\end{abstract}

\section{Introduction}
\label{sec:intro}

Reliable robot localization in operating environments is essential for their effective operation. In GPS-denied environments, this proves to be difficult. A common remedy to this problem is to localize against a prior map, often built by a previous field deployment. This approach limits localization to pre-mapped areas, and is vulnerable to scene changes over time. Additionally, if such a prebuilt map is to be built at runtime\cite{kim_scan_2018}, \cite{ORBSLAM3_TRO}, these methods scale poorly with the size of the environment. Satellite and aerial orthophotos instead provide a useful map, since they are georeferenced, widely available, and globally consistent. 

Localizing against a satellite image introduces its own set of challenges. For instance, with cameras, the gap between a top-down orthophoto and the forward camera measurement must be bridged by a view transform, termed as crossview localization~\cite{klammer_bevloc_2024}. While this gap is reduced significantly with other sensors that measure radially, such as LiDAR~\cite{vora_aerial_2020}, there is also the temporal gap from the time of capture of the satellite image. What both sensors also lack is robustness to weather, which can be bridged by using Radio Detection and Ranging (radar). Radar sensors  prove to be suitable for detection in these scenarios and yield geometric features over a longer range due to their longer wavelength~\cite{richards_principles_2010}. 

Spinning radar~\cite{noauthor_safety_nodate} presents a practical alternative to the geolocalization problem. A radar measurement is already metric and in BEV form. Existing methods that localize radar against a public map build an explicit intermediate representation first, and then solve for a geometric alignment against it using OpenStreetMap routes and semantic layers~\cite{hong_large-scale_2023}, or points extracted from overhead imagery and registered against radar-derived points~\cite{tang_rsl-net_2020},~\cite{tang_self-supervised_2020},~\cite{tang_point-based_2023},~\cite{abdullai_rascl_2025}. Each of these commits to a class label or a fixed structural representation derived from the satellite images. However, performance of such methods in unstructured environments is not explored. 

\begin{figure}[t]
\centering
\includegraphics[width=\columnwidth]{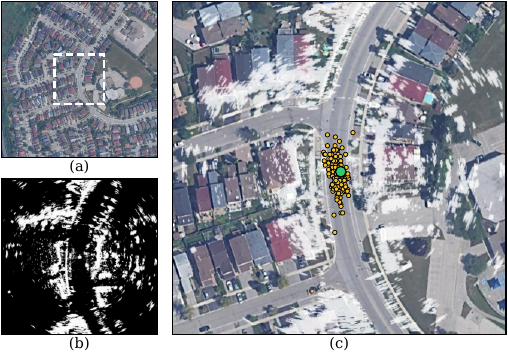}
\caption{The premise, on Boreas. \textbf{(a)} the satellite image
crop the particles are scored against, with the radar footprint boxed.
\textbf{(b)} the Cartesian radar sweep at the same instant. \textbf{(c)} The two overlaid at the filter's selected estimate from the likelihood. The structure the radar returns coincides with the structure in the image, which
is what makes them comparable. The 128 particles are drawn on top, the
one nearest ground truth in green, and the rest in yellow.}
\label{fig:teaser}
\vspace{-20pt}
\end{figure}

In this work, we instead pose a learned, shared representation for this task. A radar encoder pretrained by masked reconstruction is aligned by a contrastive objective to features of the satellite image on a shared metric grid, and the model is then asked the question: given a candidate pose, how compatible are the two representations? We use this to score pose hypotheses derived from a particle filter.  We achieve a median per-frame ATE of 2.9m and 5.17m on urban and offroad routes respectively.

Our main contributions are:
\begin{enumerate}
\item An observation model for particle filtering which computes pose likelihoods by matching learned features from a radar image and a satellite image, by explicitly learning scene structure from radar.
\item An exhaustive evaluation, to the best of our knowledge, of over 8 of 10 published datasets\footnote{We exclude the BoreasRT~\cite{lisus2026boreasroadtripmultisensor} and MOANA~\cite{jang_moana_2024} datasets, since BoreasRT provides only Doppler radar and MOANA provides marine scenarios showing very sparse returns} spanning over 31 distinct routes in both urban and offroad scenarios. 
\item An analysis of what radar reconstruction pretraining contributes, by ablating the pretrained weights.
\end{enumerate}

\section{Related Work}
\label{sec:related}

\subsection{Radar and Cross-View Geolocalisation}

The crossview geolocalization problem is well explored, primarily in computer vision. Klammer et al.~\cite{klammer_bevloc_2024} take a coarse to fine matching approach using learnt representations with a factor graph. Nash et al.~\cite{nash_censible_2024} use a modified census transform over panoramic stereo images against satellite images on the Perseverance rover to achieve sub meter global localization. Lee et al.~\cite{lee_bev-patch-pf_2026} use DINOv3\cite{simeoni_dinov3_2025} and the BEV projection coupled with a particle filter to achieve less than 8m of absolute trajectory error. 

Spinning radar became viable for motion estimation once the returns could be reduced to stable landmarks~\cite{cen_precise_2018}. Feature-based pipelines extract keypoints from the polar scan and register consecutive sweeps~\cite{cen_precise_2018, burnett_we_2021}, and radar SLAM adds loop closure over those estimates \cite{hong_radarslam_2020}. For radar sensors supplying doppler information, Le Gentil et al.~\cite{gentil_dro_2025} and~\cite{gentil_dr-pogo_2026} use doppler measurements and a direct registration method to yield sub meter accuracy in relative pose estimation.  
Within radar geolocalization, RSL-Net~\cite{tang_rsl-net_2020} align a live radar scan to satellite imagery by scoring rotated scans to estimate orientation, synthesizing a radar view, and recovering translation by cross-correlation. Its self-supervised follow-up~\cite{tang_self-supervised_2020} retains this basic pipeline while removing supervision. This is further developed into~\cite{tang_point-based_2023} overhead-image localization to radar and LiDAR through a learned point-based occupancy representation generated by ray tracing from the image center. This approach, coupled with OpenStreetMaps is exploited by Abdullai et al.~\cite{abdullai_rascl_2025}, and achieves less than 2m of ATE on the Boreas dataset~\cite{burnett_boreas_2023}, and also shows demonstrations on an unmanned surface vehicle. Considering an alternative public map, Hong et al.~\cite{hong_large-scale_2023} use OpenStreetMaps to exploit road segments and lane information to localize ground radar. 

While these methods show promising results in urban scenarios, they are not explored in unstructured environments, where prior information like OpenStreetMaps may not exist. Additionally, some of these approaches~\cite{tang_point-based_2023, tang_rsl-net_2020} also focus on single-frame evaluation, which does not show localization performance over a complete traversal. In \ref{related_work:learning representation} we explore how literature learns image representations.

\subsection{Learning image representations}
\label{related_work:learning representation}
We focus on self-supervised image representation learning in particular, since these provide a task-agnostic feature representation, which can be fine-tuned. 
Learning image representations without labels has converged on a few objectives. Masked image modeling reconstructs deliberately hidden pixels. MAE \cite{he_masked_2021} masks a large fraction of patches and
rebuilds them from the rest, SimMIM \cite{xie_simmim_2022} shows the same
idea needing only a minimal decoder, and PixMIM \cite{liu_pixmim_2023}
observes that a raw-pixel target spends capacity on high-frequency detail
and replaces it with a low-pass version of the input. Joint-embedding
methods drop reconstruction entirely and relate two views of one input in
feature space. VICReg \cite{bardes_vicreg_2022} keeps the embeddings of
different views close while variance and covariance terms prevent
collapse, and I-JEPA \cite{assran_self-supervised_2023-1} predicts the
representations of target blocks from a context block in latent space.
Contrastive methods, learning by separating a positive
from sampled negatives \cite{oord_representation_2019} also demonstrate comparable results.

\section{Methodology}
\label{sec:method}

We learn a radar representation and align it with satellite-image features to score candidate poses. A particle filter propagates these candidates with radar odometry and uses the scores as observation likelihoods.  
As Fig.~\ref{fig:teaser}(c) illustrates, a ground radar sweep and a top-down satellite image share structural information that can reasonably be assumed to persist across modalities, despite their different sensing mechanisms and acquisition times. A satellite-pretrained representation such as DINOv3 provides a broad aerial prior across environments and seasons. We therefore pretrain a radar encoder and align it with frozen satellite features on a shared metric grid.

\subsection{Problem Formulation}
\label{sec:problem}
We recursively estimate a ground vehicle's 3-DoF pose
$\mathbf{x}_t = (x_t,y_t,\theta_t) \in SE(2)$, where $(x_t,y_t)$ are east- and north-directed UTM coordinates in metres, and $\theta_t$ is the heading measured counter-clockwise from the east axis of the north-up georeferenced orthophoto $\mathcal{M}$. At time $t$, the system receives a
radar observation $z_t = \mathbf{P}_t$ and a relative odometry
estimate $\mathbf{u}_t \in SE(2)$. No GNSS is used after initialisation.

The problem is to estimate the belief
$p(\mathbf{x}_t \mid z_{1:t}, \mathbf{u}_{1:t}, \mathcal{M})$ with a particle
filter, represented by $N$ weighted samples
$\{(\mathbf{x}^i_t, w^i_t)\}_{i=1}^{N}$, each particle a discrete state
hypothesis with an associated weight. Weights are updated by the observation
likelihood,
\begin{equation}
w^i_t \propto w^i_{t-1}~ p(z_t \mid \mathbf{x}^i_t, \mathcal{M}).
\label{eq:bayes}
\end{equation}
Our work is focused on how to obtain this observation likelihood using learned radar to aerial feature alignment. We now elaborate the networks and relevant training objectives.

\subsection{Radar Representation Learning}
\label{sec:radarmae}
 We assume that the radar measurement in its cartesian form is an image. The radar measurement in its polar form is log-normalized, passed through an adaptive filter that zeroes every range bin at or below its mean and 1.5 standard deviations of its returns along the azimuth, and resampled to a Cartesian grid. We treat the resulting image as a spatial depiction of scene structure, not as a calibrated measurement of radar cross-section. Accordingly, our conclusions are limited to the \SIrange{77}{81}{GHz} band observed during training and all training and evaluation datasets use this band.

We pretrain the radar encoder as a masked autoencoder on radar cartesian images alone. We follow SimMIM \cite{xie_simmim_2022} with uniform random patch masking in the pixel space rather than the token space. We additionally use a low-pass filter applied to the reconstruction target to de-emphasize the network's focus on texture-rich details, as shown in PixMIM~\cite{liu_pixmim_2023} since reconstructing a low-pass version of the input suppresses the speckle and anisotropic noise inherent to spinning radar. 

The radar encoder being trained is a SuperPoint\cite{detone_superpoint_2018} style VGG (channels 1, 32, 64, 64, 128, 128), each stage has two $3\times3$ convolutions with batch normalization, max-pooling after the first four, giving a stride of 16. We fix the radar image size and resolution throughout our training and evaluation to 512x512px at 0.3m/px to match our satellite image resolution. The low-pass filter is a circular DFT mask at 0.25$\times$ Nyquist frequency. The radar reconstruction decoder inverts that stride in four steps (channels - 128, 64, 64, 32, 32), each step a single 3 × 3 convolution with batch normalization followed by a 2$\times$ bilinear upsample, and a final 3 × 3 convolution mapping to one channel at the input resolution. It is discarded after pretraining and only the encoder is carried forward. For training, we use the same training split as described in (Sec.\ref{sec:setup}). 

\subsection{Radar-Aerial Feature Alignment}
\label{sec:matching}
Having learned a radar-only representation, we now align it with a satellite image representation, similar in process to  \cite{lee_bev-patch-pf_2026}. Succinctly, the radar and satellite image representations, are both projected to a lower channel dimension along with a distinctiveness map to be learned via a contrastive alignment problem. While the network in \ref{sec:radarmae} serves as our base radar representation, we describe the elements of the remaining pipeline below:

\noindent\textbf{Radar decoder:} The decoder turns the encoder's
bottleneck, into the two maps the
filter needs. A $3\times3$ convolution (GroupNorm, ReLU) widens it to 256
channels, followed by 2D dropout at $p=0.1$ over whole channels. Two
$1\times1$ convolutions then form the outputs $\mathbf{G}$ and $\mathbf{C}$. The first output 32 channels, which a
GroupNorm turns into the descriptor map $\mathbf{G}_t$, $\ell_2$-normalised per cell at matching time. The second outputs the distinctiveness logit, which
a sigmoid turns into $\mathbf{C}_t$.

\noindent\textbf{Satellite branch:} We use DINOv3's \cite{simeoni_dinov3_2025}
SAT-493M ViT-L/16, which is pretrained on satellite imagery. The
map is assumed static at runtime, so DINO is run once offline over the whole
orthophoto and stored as a feature raster of shape $(1024, H/16, W/16)$. At
patch size 16 over \SI{0.3}{\metre}/px imagery each cell covers
\SI{4.8}{\metre}, matching the radar grid. Training and inference read a
window from this cache $80\times80$ cells (\SI{384}{\metre}) when
training, $100\times100$ (\SI{480}{\metre}) in the filter.

The learned component is a projection onto the shared descriptor space, which is a
dropout at $p=0.1$, a $1\times1$ convolution from 1024 to 32 channels, and a
GroupNorm, applied to the whole window once. Each candidate pose then samples
a $32\times32$ patch by bilinear interpolation on an affine grid rotated by
its heading, $\ell_2$-normalised per cell to match the radar descriptor.

\begin{figure}[t]
\centering
\includegraphics[width=\columnwidth]{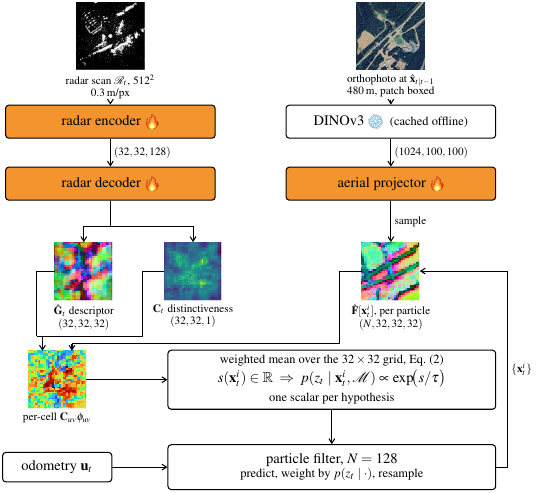}
\caption{Network architecture of the observation model.}
\vspace{-20pt}
\label{fig:network}
\end{figure}

\subsection{Training Objective}
\label{subsec:training-objective}

The training objective is designed to learn \textit{(i)} discriminative radar-aerial features, \textit{(ii)} a pixel-wise distinctiveness map for spatial weighting.

\noindent\textbf{Matching loss:}
We train the radar and aerial descriptor maps using an InfoNCE objective over
candidate aerial poses. For each radar sweep, the ground-truth pose
$\mathbf{x}^{+}$ defines the positive aerial patch. We form a negative set
$\mathcal{X}^{-}$ of 63 poses by sampling the radial offset log-uniformly,
$p(r)\propto1/r$, over \SIrange{1}{60}{\metre} about $\mathbf{x}^{+}$, and
sampling heading uniformly within $\pm~\ang{90}$ of the vehicle heading. Each candidate pose samples an aerial patch and receives the distinctiveness-weighted similarity score $s(\mathbf{x})$. For a candidate pose $\mathbf{x}$, we sample the corresponding aerial feature
patch $\hat{\mathbf{F}}[\mathbf{x}]$ and compare it with the radar descriptor map $\hat{\mathbf{G}}$. The distinctiveness-weighted similarity score is
\begin{equation}
s(\mathbf{x}) =
\frac{\sum_{u,v} \mathbf{C}_{uv}\,
      \hat{\mathbf{G}}_{uv}^{\top}
      \hat{\mathbf{F}}[\mathbf{x}]_{uv}}
     {\sum_{u,v} \mathbf{C}_{uv}}
\label{eq:score}
\end{equation}
where $\hat{\mathbf{G}}$ and $\hat{\mathbf{F}}$ are $\ell_2$-normalized feature maps.
The matching loss is
\begin{equation}
\mathcal{L}_{\mathrm{match}}
=
-\log
\frac{\exp(s(\mathbf{x}^{+})/\tau)}
{\exp(s(\mathbf{x}^{+})/\tau)
+ \sum_{\mathbf{x}\in\mathcal{X}^{-}}
\exp(s(\mathbf{x})/\tau)}.
\label{eq:match-loss}
\end{equation}
We stop gradients through the distinctiveness map when computing
$\mathcal{L}_{\mathrm{match}}$, so it is learned through its dedicated
supervision rather than being adjusted solely to reduce the matching loss.


\noindent\textbf{Distinctiveness loss:}
A low radar return is ambiguous since it may indicate that a cell contains no reflecting structure, or that the structure was weakly observed or occluded. Weighting all cells equally treats these cases identically. 
Following \cite{lee_bev-patch-pf_2026}, we
predict a distinctiveness map $\mathbf{C}$ from radar features. Let
$\phi_{uv}(\mathbf{x}) =
\hat{\mathbf{G}}_{uv}^{\top}\hat{\mathbf{F}}[\mathbf{x}]_{uv}$
denote the per-cell cosine similarity, and let
$\mathbf{x}^{-}_{\mathrm{hard}} =
\arg\max_{\mathbf{x}\in\mathcal{X}^{-}} s(\mathbf{x})$
be the hardest negative. We define the target map as
\begin{equation}
\begin{aligned}
\mathbf{C}^{*}_{uv}
&= \sigma\!\left(\phi_{uv}(\mathbf{x}^{+})\right) \\
&\quad \cdot
\sigma\!\left(\phi_{uv}(\mathbf{x}^{+})
- \phi_{uv}(\mathbf{x}^{-}_{\mathrm{hard}})\right),
\end{aligned}
\label{eq:distinctiveness-target}
\end{equation}
where $\sigma$ is the sigmoid function. We train $\mathbf{C}$ with
$\mathcal{L}_{\mathrm{distinct}} =
\operatorname{BCE}(\mathbf{C}, \mathbf{C}^{*})$ and minimize
$\mathcal{L}_{\mathrm{total}} =
\mathcal{L}_{\mathrm{match}} + \mathcal{L}_{\mathrm{distinct}}$.

\subsection{Particle Filter}
\label{sec:filter}
\paragraph{Initialisation}
$N=128$ particles are drawn from a Gaussian distribution about the first ground-truth pose,
with $\sigma = \SI{2}{\metre}$ in each horizontal axis and
$\SI{0.05}{\radian}$ ($\SI{2.9}{\degree}$) in yaw. The filter
then enters the recursive predict-update cycle and receives no further global
information.

\paragraph{Prediction step}
Each particle is propagated by the motion estimate with additive noise on the
Lie algebra,
\begin{equation}
\mathbf{x}^i_t = \mathbf{x}^i_{t-1} \oplus (\mathbf{u}_t \oplus \epsilon^i_t),
\qquad \epsilon^i_t = \mathrm{Exp}(\boldsymbol{\delta}^i_t),
\label{eq:predict}
\end{equation}
where $\oplus$ is composition on $SE(2)$ and
$\boldsymbol{\delta}^i_t \in \mathbb{R}^3$ is zero-mean Gaussian with
per-axis standard deviation proportional to $|\mathbf{u}_t|$.

\paragraph{Update step}
The network of Sec.~\ref{sec:matching} produces $\mathbf{G}_t$ and
$\mathbf{C}_t$ from $z_t$, and one projected aerial window
$\mathbf{F}_t$ about the predicted pose $\hat{\mathbf{x}}_{t|t-1}$, from
which every hypothesis samples its own patch $\mathbf{F}_t[\mathbf{x}^i_t]$. Out-of-bounds coordinates are zero-padded and masked. Because $\mathbf{F}_t$
is finite, it bounds how far a hypothesis can be scored at all. We call this bound the \emph{blind radius}, which is 163m for our work. The matching score is the distinctiveness-weighted cosine similarity as in Eq.~\ref{eq:score}. The score becomes a likelihood through a learned temperature $\tau$,
\begin{equation}
p(z_t \mid \mathbf{x}^i_t, \mathcal{M}) \propto
  \exp\!\left(s(\mathbf{x}^i_t)/\tau\right),
\label{eq:lik}
\end{equation}
and weights are updated by \eqref{eq:bayes} and normalized. $\tau$ is the
temperature learned by the contrastive objective. 
Resampling is performed whenever the effective sample size falls below half the population, with a jitter of $\SI{0.5}{\metre}$ and $\SI{0.02}{\radian}$ applied to the resampled set. The process noise of \eqref{eq:predict} is proportional to the action itself, $\sigma = (0.5, 0.5, 0.05) \odot |\mathbf{u}_t|$.
\section{Experimental Setup}
\label{sec:setup}

We hold out entire datasets instead of geographically splitting each evaluated trajectory to observe if the model can generalize across datasets. Training uses RADIATE\cite{sheeny_radiate_2021}, MulRan\cite{kim_mulran_2020},
OORD\cite{gadd_oord_2024} and GOd\cite{jiang_go_2026} (20 distinct routes). MulRan DCC03 and OORD two\_lochs are
held out for checkpoint selection alone and are reported in neither table.
Our test results are reported on Boreas~\cite{burnett_boreas_2023}, Oxford~\cite{barnes_oxford_2020}, HeRCULES~\cite{kim_hercules_2025} and FoMo~\cite{boxan_fomo_2026} (11 distinct routes), so no datasets share any route or map with training. Separately, we report results on Boreas Glen Shields and FoMo's yellow and orange trajectories over repeated traversals with varying environmental conditions to observe robustness to weather. 
\label{sec:imagery}
All maps are Google satellite imagery at 0.30m/px, exported only once per route, so that every traversal of a route is localized against an identical map. Since the particle filter can accept any odometry source for its action, we limit ourselves to a simple radar-only odometry~\cite{burnett_we_2021} without motion compensation and doppler distortion correction. This yields a radar-only result, with no GNSS after initialization. 

\subsection{Training}
\label{sec:training}

All training was performed on a system with 4x NVIDIA RTX 6000 GPUs. All evaluated results are shown on the same system, but to show realtime capabilities we show timing information on a desktop NVIDIA RTX4070 and an AMD Ryzen 7700X CPU with 64GB RAM.

\paragraph{Radar Reconstruction}
The radar encoder is pretrained as a masked autoencoder for 800 epochs with AdamW ($\text{lr}=1.5\times10^{-4}$, weight decay $0.05$, batch $128$), under a cosine schedule with a 10-epoch linear warm-up decaying to $1\times10^{-6}$. The loss is $L1$ on masked patches only, against the low-pass target of Sec.~\ref{sec:radarmae}. We use uniformly random masking, and the mask ratio ramps linearly from \SI{10}{\percent} to \SI{75}{\percent} over the first 20 epochs with each mask being a patch of size $16\times16$px.

\paragraph{Cross modal alignment}
Training runs for 50 epochs of 1000 steps each at batch $8$ with AdamW, cosine-annealed from $1\times10^{-4}$ to $1\times10^{-5}$ with no warm-up. Three parameter groups carry different rates. The projection heads (radar decoder and aerial projection) take the base rate. The radar encoder takes 0.1 times the base rate, since it arrives pretrained. The learnable InfoNCE temperature takes $0.1\times$ with weight decay disabled, learned as $\log(1/\tau)$ alongside the encoder, initialized at \num{0.05} and constrained only from below at \num{0.01}, which keeps the likelihood from sharpening without limit. We fix the temperature to 0.0387 during our evaluations, which is the learned value. For checkpoint selection, we select the checkpoint with the highest top-1 accuracy on the validation routes, evaluated on the same contrastive task used for training.
\subsection{Metrics}
We report absolute trajectory error (ATE) RMSE against post-processed
GNSS/INS ground truth over the whole trajectory, and decompose it into
along-track, cross-track and heading components in the vehicle frame. Because
mean error is dominated by rare divergence events, we additionally report the
median and the fraction of frames tracked. A sequence counts as localized if its ATE RMSE stays under
\SI{10}{\metre} in urban environments and \SI{20}{\metre} off-road. 

Given the experiments, we ask the following questions of our results:
\label{sec:questions}

\begin{itemize}
\item[\textbf{Q1}] \emph{Does our method transfer across datasets?}
  In addition to tracking accuracy, we ask if the method can perform consistently across terrain and seasons.
\item[\textbf{Q2}] \emph{Contribution of the learned radar representation.}
  To understand what explicitly learning scene structure from radar contributes, we discard the pretrained weights obtained via reconstruction.
\end{itemize}

\section{Results}
\label{sec:results}

\subsection{Performance across datasets}
\label{sec:generalisation}
\begin{table}[h]
\centering
\caption{Comparison on the Boreas sequences of \cite{abdullai_rascl_2025}.}
\label{tab:rascl}
\begin{tabular}{lS[table-format=1.2]S[table-format=1.2]S[table-format=1.2]S[table-format=1.2]S[table-format=1.2]S[table-format=1.2]S[table-format=1.2]}
\toprule
\multirow{2}{*}{Method} & \multicolumn{4}{c}{ATE RMSE} & \multicolumn{3}{c}{Decomposition} \\
\cmidrule(lr){2-5}\cmidrule(lr){6-8}
 & {01} & {02} & {03} & {04} & {Lat.} & {Long.} & {Yaw} \\
\midrule
Hong \cite{hong_large-scale_2023} & 4.70 & 4.90 & 4.90 & 3.80 & {--} & {--} & {--} \\
Li \cite{li_get_2025} & 3.00 & 3.00 & 3.40 & 3.70 & {--} & {--} & {--} \\
RaSCL \cite{abdullai_rascl_2025}$^{*}$ & \textbf{1.30} & \textbf{1.50} & \textbf{1.60} & \textbf{1.60} & \textbf{1.08} & 1.30 & 3.13 \\
\midrule
Ours  & 1.80 & 1.82 & 1.83 & 1.73 & 1.57 & \textbf{0.88} & \textbf{1.11} \\
\bottomrule
\end{tabular}
\par\vspace{5pt}\begin{minipage}{\linewidth}\footnotesize ATE RMSE (\si{\metre}) over the whole trajectory, and RMSE decomposed in the vehicle frame over all four: \emph{Lat.}\ is cross-track, \emph{Long.}\ along-track (\si{\metre}), \emph{Yaw} in \si{\degree}. $^{*}$\cite{abdullai_rascl_2025} trained on satellite imagery within the evaluated sequence. Sequence numbering follows \cite{li_get_2025}. The error decomposition for \cite{abdullai_rascl_2025} is shown only on the test location of the trajectory.\end{minipage}
\vspace{-10pt}
\end{table}

\paragraph{Comparison with prior work}
Table~\ref{tab:rascl} compares against the sequences reported
by~\cite{abdullai_rascl_2025}. Our error is lower than both published radar-to-overhead baselines on every sequence, beating both \cite{hong_large-scale_2023} and \cite{li_get_2025}. We remain behind~\cite{abdullai_rascl_2025}, which trains and tests on the Boreas dataset, whereas our method does not see this route at training time. Our heading error is roughly $2.8\times$ lower than~\cite{abdullai_rascl_2025}, and our along-track error is comparable or better. We acknowledge that our comparison to prior work is limited, since they either show single-frame evaluations~\cite{tang_rsl-net_2020}-\cite{tang_point-based_2023} and not a standalone localization solution, or their implementations were not openly available at the time of writing. We also do not compare against localization only methods like KISS-ICP~\cite{vizzo_kiss-icp_2023}, or PyCuVSLAM~\cite{korovko_cuvslam_2025} since they solve a different problem. While both of these methods can prove to be useful in our propagation step, we instead show how our method is dependent on the quality of the propagation action in Sec.~\ref{sec:odom-fail}.

\begin{table}[h]
\centering
\caption{Per-route results.}
\label{tab:full}
\scriptsize
\setlength{\tabcolsep}{2.6pt}
\begin{tabular}{@{}llS[table-format=4.2]S[table-format=4.2]S[table-format=2.2]S[table-format=2.2]S[table-format=3.2]@{}}
\toprule
Dataset & Sequence & {ATE} & {Med.} & {Lat.} & {Long.} & {Yaw} \\
\midrule
GOd (train) & morning\_run\_1 & 269.75 & 4.53 & 2.43 & 2.85 & 9.99 \\
 & route1\_morning & 20.82 & 2.59 & 3.15 & 4.52 & 6.15 \\
 & route\_1\_run\_2 & 2.51 & 1.93 & 1.18 & 2.21 & 2.47 \\
HeRCULES (test) & Bridge01 & 978.61 & 794.89 & 4.49 & 8.44 & 2.90 \\
 & Library01 & 6.98 & 3.72 & 3.75 & 5.89 & 4.25 \\
 & Mountain01 & 5.96 & 4.53 & 4.86 & 3.45 & 2.17 \\
 & ParkingLot01 & 4.05 & 4.20 & 2.64 & 3.07 & 2.90 \\
 & RiverIsland01 & 7.14 & 5.12 & 5.17 & 4.92 & 1.39 \\
 & SportsComplex01 & 3.33 & 2.90 & 2.43 & 2.27 & 1.80 \\
 & Street01 & 11.34 & 10.75 & 7.22 & 8.74 & 4.77 \\
MulRan (train) & KAIST03 & 2.17 & 1.75 & 1.35 & 1.70 & 1.08 \\
 & Riverside03 & 4.92 & 2.41 & 2.39 & 4.30 & 0.63 \\
 & Sejong03 & 616.87 & 358.71 & 5.98 & 4.50 & 1.10 \\
OORD (train) & 2021-11-25$^{\dag}$ & 1525.74 & 1510.82 & 3.92 & 7.65 & 6.97 \\
 & 2021-11-27$^{\dag}$ & 1885.71 & 1358.48 & 4.72 & 8.23 & 9.47 \\
 & 2021-11-28 & 15.90 & 5.41 & 5.79 & 5.92 & 7.07 \\
Oxford (test) & 2019-01-14 & 7.39 & 3.96 & 5.22 & 5.24 & 3.07 \\
RADIATE (train) & city\_1\_1$^{\dag}$ & 487.19 & 153.18 & 3.42 & 16.70 & 15.88 \\
 & city\_1\_3 & 11.88 & 6.83 & 1.06 & 11.41 & 1.78 \\
 & city\_2\_0 & 8.02 & 4.78 & 2.36 & 7.66 & 6.87 \\
 & city\_3\_0 & 6.66 & 3.33 & 1.96 & 6.37 & 5.84 \\
 & city\_3\_1 & 7.17 & 6.11 & 1.60 & 6.99 & 1.84 \\
 & city\_5\_0 & 8.33 & 4.89 & 2.19 & 8.04 & 4.19 \\
 & city\_6\_0 & 19.26 & 14.72 & 1.96 & 17.01 & 6.70 \\
 & motorway\_1\_0 & 15.43 & 13.33 & 3.98 & 14.28 & 11.26 \\
 & rain\_4\_1 & 8.61 & 5.50 & 1.66 & 8.45 & 6.30 \\
 & rural\_1\_1$^{\dag}$ & 1147.61 & 1281.18 & 1.84 & 12.50 & 6.94 \\
 & rural\_1\_3 & 12.03 & 11.48 & 2.04 & 11.86 & 5.92 \\
\midrule
\multicolumn{2}{l}{\textbf{median}} & 9.98 & 5.26 & 2.54 & 6.68 & 4.51 \\
\bottomrule
\end{tabular}
\par\smallskip
\parbox{\columnwidth}{\footnotesize \emph{Lat.}\ is cross-track and \emph{Long.}\ along-track, RMS in the body frame. $^{\dag}$Re-anchoring at ground truth once the degeneracy ends yield an ATE RMSE of \SI{9.62}{\metre}, \SI{8.45}{\metre}, \SI{98.91}{\metre}, \SI{23.82}{\metre} respectively. On OORD, the scan matcher degrades over featureless terrain, returning a confident near-zero translation; on RADIATE the published ground truth is itself unreliable, with 89 of 720 frames stepping backwards by up to \SI{12}{\metre} and one \SI{79}{\metre} discontinuity between consecutive frames.}
\vspace{-10pt}
\end{table}

\begin{figure}[t]
\centering
\includegraphics[width=\columnwidth]{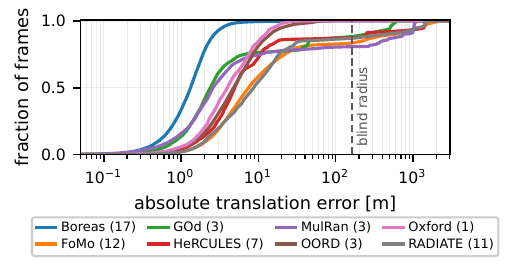}
\caption{Cumulative distribution of per-frame position error, by dataset.
The dashed line is the blind radius; a dataset whose curve reaches 1.0 to the left
of it never lost lock.}
\label{fig:error-cdf}
\vspace{-10pt}
\end{figure}

\paragraph{Test Datasets}
Tables~\ref{tab:full}--\ref{tab:temporal} collectively report every
route and traversal in the test set. We observe that accuracy when tracking is nearly
environment-independent(Fig.~\ref{fig:error-cdf}). In urban scenarios, median ATE lies within \SIrange{1.0}{4.7}{\metre} and off-road scenarios at \SIrange{5.7}{9.6}{\metre}. The Boreas traversals (Tab.~\ref{tab:boreas-all}) spanning a full year on one map lie within \SIrange{1.19}{1.58}{\metre} median ATE. 

We also observe that localization failures are distinct events and not gradual drift. Each sequence either tracks at $<10$m error or diverges past the blind radius and is lost. The gap in performance between offroad and urban scenarios is not degraded accuracy but a small number of divergence events, which we discuss in Sec.~\ref{sec:disc-q2}.

\begin{figure*}[t]
\centering
\includegraphics[width=\textwidth]{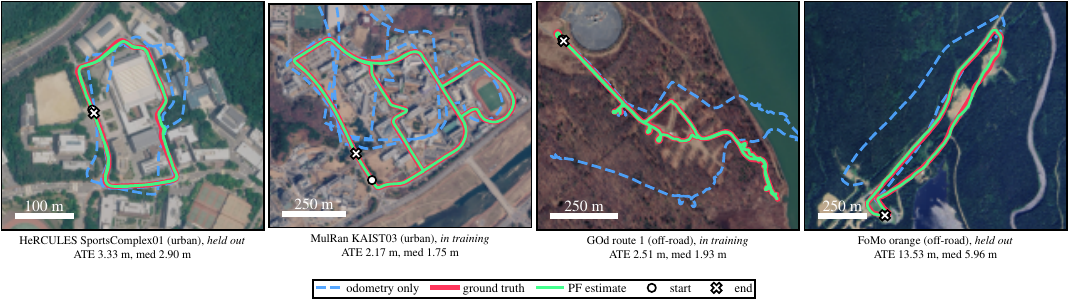}
\caption{Estimated trajectories on four routes spanning four datasets, plotted with ground truth and the scan-matching
odometry the filter propagates on. While the odometry drifts on every route, the filter estimate stays on ground truth, showing that the observation model provides meaningful corrections. }
\label{fig:traj-qualitative}
\vspace{-20pt}
\end{figure*}

\paragraph{Train Datasets}
Table~\ref{tab:full} reports the routes within the training set, evaluated
against the same maps. 

9 of 20 routes in training meet the pass threshold, against 9 of 11 routes (20 of 26 traversals) from the held-out datasets. This gap in the pass rate stems from the composition of the training datasets. For the OORD and RADIATE datasets in particular, we argue they are genuinely challenging either due to lack of matchable structure for the propagation source, or due to data drops. The two OORD routes that fail at over \SI{900}{\metre} track at \SI{4.4}{}--\SI{5.2}{\metre} median while the odometry works, so their error is due to odometry divergence rather than degraded matching.

\subsection{Robustness over a year}
\label{sec:temporal}

\begin{table}[t]
\centering
\caption{Boreas: one route, thirteen months. One traversal for every unique weather condition the dataset labels (14 of them).}
\label{tab:boreas-all}
\setlength{\tabcolsep}{2.5pt}
\scriptsize
\begin{tabular}{llS[table-format=2.2]S[table-format=1.2]S[table-format=2.2]S[table-format=1.2]S[table-format=3.2]}
\toprule
Date & Conditions & {ATE} & {Median} & {Lat.} & {Long.} & {Yaw} \\
 & & {[\si{\metre}]} & {[\si{\metre}]} & {[\si{\metre}]} & {[\si{\metre}]} & {[\si{\degree}]} \\
\midrule
2020-12-18 & Sun, Snow & 2.68 & 1.34 & 2.51 & 0.93 & 1.34 \\
2021-01-15 & Sun, Clouds, Snow & 1.98 & 1.34 & 1.77 & 0.88 & 1.25 \\
2021-01-19 & Clouds, Snow & 1.97 & 1.25 & 1.78 & 0.84 & 1.34 \\
2021-01-26 & Overcast, Snow, Snowing & 2.27 & 1.58 & 1.92 & 1.21 & 1.35 \\
2021-02-02 & Overcast, Snow & 2.02 & 1.35 & 1.81 & 0.90 & 1.31 \\
2021-03-23$^{*}$ & Overcast, Construction & 2.09 & 1.37 & 1.89 & 0.90 & 178.92 \\
2021-04-08 & Sun & 2.11 & 1.38 & 1.94 & 0.83 & 1.36 \\
2021-04-13 & Sun, Clouds, Construction & 2.24 & 1.38 & 2.08 & 0.86 & 1.39 \\
2021-04-20 & Clouds, Construction & 1.82 & 1.26 & 1.60 & 0.87 & 1.36 \\
2021-04-29 & Overcast, Rain & 1.99 & 1.36 & 1.75 & 0.94 & 1.51 \\
2021-05-06 & Sun, Clouds & 2.01 & 1.32 & 1.79 & 0.92 & 1.37 \\
2021-09-14 & Night & 2.43 & 1.31 & 2.28 & 0.84 & 1.45 \\
2021-10-22 & Clouds & 2.23 & 1.25 & 2.02 & 0.95 & 1.20 \\
2021-11-14 & Overcast & 1.82 & 1.19 & 1.61 & 0.85 & 1.22 \\
\midrule
\textbf{median} & & 2.06 & 1.34 & 1.85 & 0.89 & 1.35 \\
\bottomrule
\end{tabular}
\par\smallskip
\parbox{\columnwidth}{\footnotesize Glen Shields traversals against one satellite tile. \emph{Lat.}\ is cross-track and \emph{Long.}\ along-track, RMS in the body frame over all frames. $^{*}$This traversal's published ground-truth heading points \SI{180}{\degree} from its direction of travel, so a filter seeded from it propagates backwards along the route from the first frame; the yaw column is measured against the published heading.}
\vspace{-20pt}
\end{table}
\begin{table}[t]
\centering
\caption{FoMo: two off-road routes, six traversals each over a year.}
\label{tab:temporal}
\footnotesize
\setlength{\tabcolsep}{3.5pt}
\begin{tabular}{@{}lS[table-format=4.0]S[table-format=1.1]S[table-format=3.0]S[table-format=+2.1]S[table-format=4.2]S[table-format=4.2]@{}}
\toprule
Date & {Snow} & {Rain} & {RH} & {Temp.} & {ATE} & {Median} \\
 & {[\si{\milli\metre}]} & {[\si{\milli\metre}]} & {[\si{\percent}]} & {[\si{\celsius}]} & {[\si{\metre}]} & {[\si{\metre}]} \\
\midrule
\multicolumn{7}{@{}l}{\emph{yellow route}} \\
2024-11-28 & {--} & 0.0 & 89 & -2.5 & 13.20 & 5.96 \\
2025-03-10 & 1060 & 0.0 & 79 & -7.1 & 84.70 & 7.60 \\
2025-08-20 & 0 & 0.0 & 45 & +18.3 & 10.47 & 4.90 \\
2025-09-24 & 0 & 0.0 & 50 & +16.2 & 9.00 & 5.32 \\
2025-10-14 & 3 & 0.0 & 40 & +17.1 & 8.30 & 4.47 \\
2025-11-03 & 0 & 0.4 & 100 & +1.6 & 7.70 & 3.83 \\
\addlinespace[2pt]\midrule
\multicolumn{7}{@{}l}{\emph{orange route}} \\
2025-01-10 & 502 & 0.0 & 72 & -7.7 & 714.11 & 228.59 \\
2025-03-14 & 1038 & 0.0 & 55 & +0.4 & 473.71 & 54.89 \\
2025-08-20 & 0 & 0.0 & 44 & +19.4 & 13.08 & 7.57 \\
2025-09-24 & 0 & 0.0 & 51 & +16.0 & 11.86 & 5.23 \\
2025-10-14 & 1 & 0.0 & 38 & +15.4 & 13.53 & 5.96 \\
2025-11-03$^{\dag}$ & 0 & 0.6 & 100 & +1.5 & 1213.62 & 1000.71 \\
\bottomrule
\end{tabular}
\par\smallskip
\parbox{\columnwidth}{\footnotesize Conditions reported from \cite{boxan_fomo_2026}. Depths below the sensor's zero are reported as \SI{0}{\milli\metre}, and a dash marks a traversal with no snow record. $^{\dag}$ This traversal fails for a reason unrelated to conditions, the radar odometry accumulates \SI{211}{\degree} of yaw drift over the route, which carries the filter outside the blind radius.}
\vspace{-10pt}
\end{table}

We evaluate on every unique environmental condition on the Glen Shields route of the Boreas dataset, over thirteen
months. Here, the route, map, and model are fixed, so only the radar measurement changes over time. Table~\ref{tab:boreas-all} gives one
traversal for each weather condition using the dataset labels. All fourteen fall between \SI{1.82}{} and \SI{2.68}{\metre}, with a median per-frame error of \SI{1.34}{\metre}. We observe no measurable degradation between sun and snow conditions, which demonstrates the robustness of radar as a sensor expressing structure than our method. 

Table~\ref{tab:temporal} pairs the same question with the FoMo dataset's \emph{Orange} and \emph{Yellow} routes, where snow depth, rainfall, humidity, and temperature are available from the dataset.  Across both routes every traversal localizes, at \SIrange{3.8}{7.6}{\metre} median, and all three traversals over half a meter or more of snow do not. We argue this is because accumulated snow changes the scene structure enough to show a difference between the available radar measurement and the satellite image. A satellite image captured around the same time as the dataset was not available to test further. 

\subsection{Contribution of the learned radar representation}
\label{sec:pretrain}
\begin{table}[t]
\centering
\caption{Encoder initialisation. Median ATE over each dataset's traversals, and the number that lose lock. Held-out datasets above the rule, training datasets below.}
\label{tab:pretrain}
\footnotesize
\begin{tabular}{lS[table-format=2.0]S[table-format=4.2]cS[table-format=4.2]c}
\toprule
& & \multicolumn{2}{c}{MAE init} & \multicolumn{2}{c}{Random init} \\
\cmidrule(lr){3-4}\cmidrule(lr){5-6}
Dataset & {$n$} & {ATE} & Lost & {ATE} & Lost \\
 & & {[\si{\metre}]} & & {[\si{\metre}]} & \\
\midrule
Boreas (Glen Shields) & 14 & \bfseries 2.06 & 0/14 & 1442.66 & 14/14 \\
FoMo & 12 & 13.14 & 4/12 & \bfseries 12.43 & 1/12 \\
HeRCULES & 7 & \bfseries 6.98 & 1/7 & 259.36 & 6/7 \\
Oxford & 1 & \bfseries 7.39 & 0/1 & 656.32 & 1/1 \\
\midrule
MulRan & 3 & \bfseries 4.92 & 1/3 & 918.98 & 3/3 \\
RADIATE & 11 & \bfseries 11.88 & 2/11 & 279.23 & 11/11 \\
OORD & 3 & 1525.74 & 2/3 & \bfseries 594.01 & 3/3 \\
GO & 3 & 20.82 & 1/3 & \bfseries 2.61 & 1/3 \\
\midrule
\textit{all} & 54 & & 11/54 & & 40/54 \\
\bottomrule
\end{tabular}
\vspace{-10pt}
\end{table}

Our observation model compares two learned representations, which brings the
question of what the radar reconstruction pretraining contributes, and if the contrastive objective alone is enough. We answer it by discarding the radar encoder's pretrained weights. The network is trained with the radar encoder initialized randomly, and everything else is fixed (Table~\ref{tab:pretrain}). On the datasets held out of training, initialization contributes substantially, with Boreas going from a median
\SI{2.06}{\metre} to diverging on all \num{14} traversals, and on the HeRCULES dataset
from one failure in seven to six. The training datasets behave similarly,
which discards a generalization gap as the explanation. MulRan, RADIATE and
OORD all fail on routes the randomly initialized model was trained on. The
contrastive objective can align an encoder that already represents radar
structure to overhead features, but cannot learn that structure from scratch in this training duration. Across all \num{54} sequences, \num{11} lose lock with pretraining and \num{40} without. On the two off-road routes which improve without the pretraining, we argue that the present scene structure is sparse enough to be learned within the contrastive training. On GO, the site with the
sparsest structure, the randomly initialized encoder reaches a median of
\SI{2.61}{\metre} against \SI{20.82}{\metre}. On OORD, the median ATE is lower but fragile, since it drifts more often. 
\vspace{-5pt}
\subsection{Runtime}
\label{sec:timing}
Table~\ref{tab:timing} reports the per-frame duration of the localization stage on
a desktop RTX 4070. The observation model runs at \SI{32.8}{\hertz}, eight times the \SI{4}{\hertz} sensor rate. The learned components are \SI{11.5}{\percent} of that frame. End to end, including radar preprocessing and scan-matching odometry, the pipeline
runs at \SI{6.23}{\hertz}, with odometry alone occupying \SI{48}{\percent} of the duration.
\vspace{-10pt}
\begin{table}[h]
\centering
\caption{Per-frame runtime of the localization stage.}
\label{tab:timing}
\begin{tabular}{lS[table-format=2.2]S[table-format=3.1]}
\toprule
Stage & {Time [\si{\milli\second}]} & {Share [\si{\percent}]} \\
\midrule
Aerial feature crop & 24.73 & 81.2 \\
Host-to-device transfer & 2.12 & 7.0 \\
Radar encoder & 2.38 & 7.8 \\
Aerial projection & 0.26 & 0.9 \\
Scoring, 128 particles & 0.86 & 2.8 \\
Device-to-host, filter update & 0.12 & 0.4 \\
\midrule
Total & 30.47 & 100.0 \\
\bottomrule
\end{tabular}
\par\smallskip
\begin{minipage}{\linewidth}\footnotesize Median over all frames computed on the Boreas Dataset. \end{minipage}
\vspace{-20pt}
\end{table}

\section{Discussion}
\label{sec:discussion}

Given our results, we observe that our method is able to generalize across datasets. We attribute this to what the objective learns and not the radar representation itself. Moreover, DINOv3 provides a strong visual prior. So the generality is inherited and not acquired from the training datasets. While most of the routes in the training set show failures from the odometry, this can be verified by using a different odometry source.

\subsection{Failure Modes}
\label{sec:disc-q2}
\label{sec:odom-fail}
We now analyze what causes our method to fail.

\paragraph{Odometry-driven divergence}

The most frequent cause is the propagation source. HeRCULES
Bridge01 (row 4 of Tab.~\ref{tab:full}) presents the worst result in yhr HeRCULES routes at \SI{979}{\metre}, with
\SI{88.7}{\percent} of total frames outside the blind radius. This is due to degraded geometry as the Bridge01 route lies on a bridge, and the radar measurement only observes the edges shown by the dividers as straight lines. This is analogous to LiDAR and camera odometry pipelines failing in textureless tunnels. Yet when the odometry does perform well, the matching score from the observation model is higher than that of SportsComplex01(row 9 of Tab.~\ref{tab:full}), which localizes at \SI{3.33}{\metre} ATE. We find the same diagnosis for the two OORD routes which have an ATE RMSE above \SI{900}{\metre} in Table~\ref{tab:full}, which track at \SI{4.4}{}-\SI{5.2}{\metre} respectively before odometry degrades. This shows that our method is dependent on the quality of the propogation source. 

\paragraph{Uninformative-likelihood divergence}
We find this to be the dominant failure mode in offroad environments. While tracking, the median ATE on FoMo yellow (rows 2025-08-20 and 2025-03-10 in Table~\ref{tab:temporal}) moves from 4.9m to 7.6m, yet the ATE RMSE from \SI{10.47}{\metre} to \SI{84.70}{\metre}. Since shared structure between the radar and the satellite image is sparse, the observation likelihood flattens. This drives the filter estimate outside the blind radius, so recovery from such a divergence is not possible. This is also what separates the off-road rows of Table~\ref{tab:full} from the urban ones. We hypothesize that this is a function of the terrain differing from the satellite map we use. A direct example of this is the 2025-03-10 traversal from FoMo's Yellow route, where snow accumulates over a height of 1m (Tab.~\ref{tab:temporal}). Radar is robust to snowfall to a degree, but accumulation of this depth is consistent with a loss of the structure the two views share. Shared structure between the two views can also be reduced due to heavy tilt in the trajectory, as is shown on the Orange trajectory in FoMo (last row in Tab.~\ref{tab:temporal}), also shown by \cite{boxan_radar_2026}

\paragraph{Orthorectification}
\begin{figure}[t]
\centering
\includegraphics[width=\columnwidth]{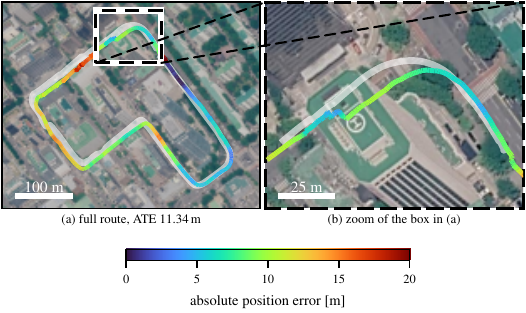}
\caption{HeRCULES Street01, with the estimate coloured by per-frame error and the
dashed box marking the zoom. Ground truth and the estimate run parallel,
separated by a sustained offset toward the buildings.}
\label{fig:ortho-lean}
\vspace{-15pt}
\end{figure}
Incorrect orthorectification causes tall objects like highrise buildings to overlap onto structures behind them. This is observed with the HeRCULES Street01(Fig.~\ref{fig:ortho-lean} and the Oxford satellite images. All \num{4180}  frames of Street01 are tracked, yet it reaches \SI{11.34}{\metre} ATE RMSE. Imagery that is not true-orthorectified moves tall structures from their footprint, while the radar observes at the ground level. This contributes to lack of shared structure, but is limited only to areas where incorrect orthorectification exists, which is observable in the inset in Fig.~\ref{fig:ortho-lean}.
Oxford shows the same mechanism. It is our weakest urban result at \SI{7.39}{\metre} within the set that passes the $<10$m ATE threshold, yet the filter never loses tracking. Abdullai et al.~\cite{abdullai_rascl_2025} reach the same conclusion on the same dataset.  
\section{Limitations and Conclusion}
\label{sec:conclusion}
Since the estimation framework used is a particle filter, the observation model only provides ranking on the pose hypotheses presented by the filter. If none of the poses correspond to the ground truth, the observation model merely picks the closest in similarity, which is insufficient to drive the estimate back to the correct location. Additionally, since the limitation is from the pose hypotheses being restricted, the observation model should instead provide a measurement independent of the pose hypotheses, as per-frame estimate methods like \cite{tang_rsl-net_2020} provide. Such an independent measurement would also alleviate the dependence on the quality of the propogation source as shown in Sec.~\ref{sec:disc-q2}.
In conclusion, we present RadGeoLoc, a particle-filter based approach for cross-view geolocalization using learned representations from both radar and satellite imagery. Owing to the radar's robustness to weather and the strong visual prior provided by DINOv3, we show results that generalize across datasets. Future work would involve a more tightly coupled approach using a fixed-lag smoother to revise previous states and a global relocalization method to provide measurements independent of a pose hypothesis. 

\bstctlcite{IEEEexample:BSTcontrol}
{\footnotesize
\bibliographystyle{IEEEtran}
\bibliography{IEEEabrv, references, references-2}

\begin{thebibliography}{10}
\providecommand{\url}[1]{#1}
\csname url@rmstyle\endcsname
\providecommand{\newblock}{\relax}
\providecommand{\bibinfo}[2]{#2}
\providecommand\BIBentrySTDinterwordspacing{\spaceskip=0pt\relax}
\providecommand\BIBentryALTinterwordstretchfactor{4}
\providecommand\BIBentryALTinterwordspacing{\spaceskip=\fontdimen2\font plus
\BIBentryALTinterwordstretchfactor\fontdimen3\font minus \fontdimen4\font\relax}
\providecommand\BIBforeignlanguage[2]{{%
\expandafter\ifx\csname l@#1\endcsname\relax
\typeout{** WARNING: IEEEtran.bst: No hyphenation pattern has been}%
\typeout{** loaded for the language `#1'. Using the pattern for}%
\typeout{** the default language instead.}%
\else
\language=\csname l@#1\endcsname
\fi
#2}}

\bibitem{kim_scan_2018}
\BIBentryALTinterwordspacing
G.~Kim and A.~Kim, ``Scan {Context}: {Egocentric} {Spatial} {Descriptor} for {Place} {Recognition} {Within} {3D} {Point} {Cloud} {Map},'' in \emph{2018 {IEEE}/{RSJ} {International} {Conference} on {Intelligent} {Robots} and {Systems} ({IROS})}, Oct. 2018, pp. 4802--4809, iSSN: 2153-0866. [Online]. Available: \url{https://ieeexplore.ieee.org/document/8593953}
\BIBentrySTDinterwordspacing

\bibitem{ORBSLAM3_TRO}
C.~Campos, R.~Elvira, J.~J.~G. Rodríguez, J.~M. M.~Montiel, and J.~D.~Tardós, ``Orb-slam3: An accurate open-source library for visual, visual–inertial, and multimap slam,'' \emph{IEEE Transactions on Robotics}, vol.~37, no.~6, pp. 1874--1890, 2021.

\bibitem{klammer_bevloc_2024}
\BIBentryALTinterwordspacing
C.~Klammer and M.~Kaess, ``{BEVLoc}: {Cross}-{View} {Localization} and {Matching} via {Birds}-{Eye}-{View} {Synthesis},'' Oct. 2024, arXiv:2410.06410 [cs.RO]. [Online]. Available: \url{http://arxiv.org/abs/2410.06410}
\BIBentrySTDinterwordspacing

\bibitem{vora_aerial_2020}
\BIBentryALTinterwordspacing
A.~Vora, S.~Agarwal, G.~Pandey, and J.~McBride, ``Aerial {Imagery} based {LIDAR} {Localization} for {Autonomous} {Vehicles},'' Mar. 2020, arXiv:2003.11192 [cs.RO]. [Online]. Available: \url{http://arxiv.org/abs/2003.11192}
\BIBentrySTDinterwordspacing

\bibitem{richards_principles_2010}
\BIBentryALTinterwordspacing
M.~A. Richards, J.~A. Scheer, and W.~A. Holm, \emph{Principles of {Modern} {Radar}, {Volume} {I} - {Basic} {Principles}}.\hskip 1em plus 0.5em minus 0.4em\relax SciTech Publishing, 2010. [Online]. Available: \url{https://app.knovel.com/hotlink/toc/id:kpPMRVIBP8/principles-modern-radar/principles-modern-radar}
\BIBentrySTDinterwordspacing

\bibitem{noauthor_safety_nodate}
\BIBentryALTinterwordspacing
``\BIBforeignlanguage{en}{Safety {\textbar} {Security} {\textbar} {Efficiency}}.'' [Online]. Available: \url{https://navtechradar.com/}
\BIBentrySTDinterwordspacing

\bibitem{hong_large-scale_2023}
\BIBentryALTinterwordspacing
Z.~Hong, Y.~Petillot, K.~Zhang, S.~Xu, and S.~Wang, ``Large-{Scale} {Radar} {Localization} using {Online} {Public} {Maps},'' in \emph{2023 {IEEE} {International} {Conference} on {Robotics} and {Automation} ({ICRA})}, May 2023, pp. 3990--3996. [Online]. Available: \url{https://ieeexplore.ieee.org/document/10160730}
\BIBentrySTDinterwordspacing

\bibitem{tang_rsl-net_2020}
\BIBentryALTinterwordspacing
T.~Y. Tang, D.~De~Martini, D.~Barnes, and P.~Newman, ``\BIBforeignlanguage{en}{{RSL}-{Net}: {Localising} in {Satellite} {Images} {From} a {Radar} on the {Ground}},'' \emph{\BIBforeignlanguage{en}{IEEE Robotics and Automation Letters}}, vol.~5, no.~2, pp. 1087--1094, Apr. 2020. [Online]. Available: \url{https://ieeexplore.ieee.org/document/8957240/}
\BIBentrySTDinterwordspacing

\bibitem{tang_self-supervised_2020}
\BIBentryALTinterwordspacing
T.~Tang, D.~De~Martini, S.~Wu, and P.~Newman, ``\BIBforeignlanguage{en}{Self-{Supervised} {Localisation} between {Range} {Sensors} and {Overhead} {Imagery}},'' in \emph{\BIBforeignlanguage{en}{Robotics: {Science} and {Systems} {XVI}}}.\hskip 1em plus 0.5em minus 0.4em\relax Robotics: Science and Systems Foundation, July 2020. [Online]. Available: \url{http://www.roboticsproceedings.org/rss16/p057.pdf}
\BIBentrySTDinterwordspacing

\bibitem{tang_point-based_2023}
\BIBentryALTinterwordspacing
T.~Y. Tang, D.~De~Martini, and P.~Newman, ``\BIBforeignlanguage{en}{Point-based metric and topological localisation between lidar and overhead imagery},'' \emph{\BIBforeignlanguage{en}{Autonomous Robots}}, vol.~47, no.~5, pp. 595--615, June 2023. [Online]. Available: \url{https://doi.org/10.1007/s10514-023-10085-w}
\BIBentrySTDinterwordspacing

\bibitem{abdullai_rascl_2025}
\BIBentryALTinterwordspacing
B.~Abdullai, T.~Wang, X.~Qiao, F.~Shkurti, and T.~D. Barfoot, ``{RaSCL}: {Radar} to {Satellite} {Crossview} {Localization},'' Apr. 2025, arXiv:2504.15899 [cs.RO]. [Online]. Available: \url{http://arxiv.org/abs/2504.15899}
\BIBentrySTDinterwordspacing

\bibitem{lisus2026boreasroadtripmultisensor}
\BIBentryALTinterwordspacing
D.~Lisus, K.~M. Papais, C.~L. Gentil, E.~Preston-Krebs, A.~Lambert, K.~Y.~K. Leung, and T.~D. Barfoot, ``Boreas road trip: A multi-sensor autonomous driving dataset on challenging roads,'' 2026. [Online]. Available: \url{https://arxiv.org/abs/2602.16870}
\BIBentrySTDinterwordspacing

\bibitem{jang_moana_2024}
\BIBentryALTinterwordspacing
H.~Jang, W.~Yang, H.~Kim, D.~Lee, Y.~Kim, J.~Park, M.~Jeon, J.~Koh, Y.~Kang, M.~Jung, S.~Jung, C.~Z. Hao, W.~Y. Hin, C.~Yihang, and A.~Kim, ``\BIBforeignlanguage{en}{{MOANA}: {Multi}-{Radar} {Dataset} for {Maritime} {Odometry} and {Autonomous} {Navigation} {Application}},'' Dec. 2024. [Online]. Available: \url{https://arxiv.org/abs/2412.03887v4}
\BIBentrySTDinterwordspacing

\bibitem{nash_censible_2024}
\BIBentryALTinterwordspacing
J.~Nash, Q.~Dwight, L.~Saldyt, H.~Wang, S.~Myint, A.~Ansar, and V.~Verma, ``Censible: {A} {Robust} and {Practical} {Global} {Localization} {Framework} for {Planetary} {Surface} {Missions},'' in \emph{2024 {IEEE} {International} {Conference} on {Robotics} and {Automation} ({ICRA})}, May 2024, pp. 8642--8648. [Online]. Available: \url{https://ieeexplore.ieee.org/document/10611697}
\BIBentrySTDinterwordspacing

\bibitem{lee_bev-patch-pf_2026}
\BIBentryALTinterwordspacing
D.~Lee, J.~Quattrociocchi, C.~Ellis, R.~Rana, A.~Adkins, A.~Uccello, G.~Warnell, and J.~Biswas, ``{BEV}-{Patch}-{PF}: {Particle} {Filtering} with {BEV}-{Aerial} {Feature} {Matching} for {Off}-{Road} {Geo}-{Localization},'' Mar. 2026, arXiv:2512.15111 [cs.RO]. [Online]. Available: \url{http://arxiv.org/abs/2512.15111}
\BIBentrySTDinterwordspacing

\bibitem{simeoni_dinov3_2025}
\BIBentryALTinterwordspacing
O.~Siméoni, H.~V. Vo, M.~Seitzer, F.~Baldassarre, M.~Oquab, C.~Jose, V.~Khalidov, M.~Szafraniec, S.~Yi, M.~Ramamonjisoa, F.~Massa, D.~Haziza, L.~Wehrstedt, J.~Wang, T.~Darcet, T.~Moutakanni, L.~Sentana, C.~Roberts, A.~Vedaldi, J.~Tolan, J.~Brandt, C.~Couprie, J.~Mairal, H.~Jégou, P.~Labatut, and P.~Bojanowski, ``{DINOv3},'' Aug. 2025, arXiv:2508.10104 [cs.CV]. [Online]. Available: \url{http://arxiv.org/abs/2508.10104}
\BIBentrySTDinterwordspacing

\bibitem{cen_precise_2018}
\BIBentryALTinterwordspacing
S.~H. Cen and P.~Newman, ``Precise {Ego}-{Motion} {Estimation} with {Millimeter}-{Wave} {Radar} {Under} {Diverse} and {Challenging} {Conditions},'' in \emph{2018 {IEEE} {International} {Conference} on {Robotics} and {Automation} ({ICRA})}, May 2018, pp. 6045--6052, iSSN: 2577-087X. [Online]. Available: \url{https://ieeexplore.ieee.org/document/8460687}
\BIBentrySTDinterwordspacing

\bibitem{burnett_we_2021}
\BIBentryALTinterwordspacing
K.~Burnett, A.~P. Schoellig, and T.~D. Barfoot, ``Do {We} {Need} to {Compensate} for {Motion} {Distortion} and {Doppler} {Effects} in {Spinning} {Radar} {Navigation}?'' Jan. 2021, arXiv:2011.03512 [cs.RO]. [Online]. Available: \url{http://arxiv.org/abs/2011.03512}
\BIBentrySTDinterwordspacing

\bibitem{hong_radarslam_2020}
\BIBentryALTinterwordspacing
Z.~Hong, Y.~Petillot, and S.~Wang, ``{RadarSLAM}: {Radar} based {Large}-{Scale} {SLAM} in {All} {Weathers},'' May 2020, arXiv:2005.02198 [cs.RO]. [Online]. Available: \url{http://arxiv.org/abs/2005.02198}
\BIBentrySTDinterwordspacing

\bibitem{gentil_dro_2025}
\BIBentryALTinterwordspacing
C.~L. Gentil, L.~Brizi, D.~Lisus, X.~Qiao, G.~Grisetti, and T.~D. Barfoot, ``{DRO}: {Doppler}-{Aware} {Direct} {Radar} {Odometry},'' Apr. 2025, arXiv:2504.20339 [cs.RO]. [Online]. Available: \url{http://arxiv.org/abs/2504.20339}
\BIBentrySTDinterwordspacing

\bibitem{gentil_dr-pogo_2026}
\BIBentryALTinterwordspacing
C.~L. Gentil, W.~Li, L.~Brizi, and T.~D. Barfoot, ``Dr-{PoGO}: {Direct} {Radar} {Pose}-{Graph} {Optimization},'' May 2026, arXiv:2605.04806 [cs.RO]. [Online]. Available: \url{http://arxiv.org/abs/2605.04806}
\BIBentrySTDinterwordspacing

\bibitem{burnett_boreas_2023}
\BIBentryALTinterwordspacing
K.~Burnett, D.~J. Yoon, Y.~Wu, A.~Z. Li, H.~Zhang, S.~Lu, J.~Qian, W.-K. Tseng, A.~Lambert, K.~Y.~K. Leung, A.~P. Schoellig, and T.~D. Barfoot, ``Boreas: {A} {Multi}-{Season} {Autonomous} {Driving} {Dataset},'' Jan. 2023, arXiv:2203.10168 [cs.RO]. [Online]. Available: \url{http://arxiv.org/abs/2203.10168}
\BIBentrySTDinterwordspacing

\bibitem{he_masked_2021}
\BIBentryALTinterwordspacing
K.~He, X.~Chen, S.~Xie, Y.~Li, P.~Dollár, and R.~Girshick, ``Masked {Autoencoders} {Are} {Scalable} {Vision} {Learners},'' Dec. 2021, arXiv:2111.06377 [cs.CV]. [Online]. Available: \url{http://arxiv.org/abs/2111.06377}
\BIBentrySTDinterwordspacing

\bibitem{xie_simmim_2022}
\BIBentryALTinterwordspacing
Z.~Xie, Z.~Zhang, Y.~Cao, Y.~Lin, J.~Bao, Z.~Yao, Q.~Dai, and H.~Hu, ``{SimMIM}: {A} {Simple} {Framework} for {Masked} {Image} {Modeling},'' Apr. 2022, arXiv:2111.09886 [cs.CV]. [Online]. Available: \url{http://arxiv.org/abs/2111.09886}
\BIBentrySTDinterwordspacing

\bibitem{liu_pixmim_2023}
\BIBentryALTinterwordspacing
Y.~Liu, S.~Zhang, J.~Chen, K.~Chen, and D.~Lin, ``{PixMIM}: {Rethinking} {Pixel} {Reconstruction} in {Masked} {Image} {Modeling},'' Mar. 2023, arXiv:2303.02416 [cs.CV]. [Online]. Available: \url{http://arxiv.org/abs/2303.02416}
\BIBentrySTDinterwordspacing

\bibitem{bardes_vicreg_2022}
\BIBentryALTinterwordspacing
A.~Bardes, J.~Ponce, and Y.~LeCun, ``{VICReg}: {Variance}-{Invariance}-{Covariance} {Regularization} for {Self}-{Supervised} {Learning},'' Jan. 2022, arXiv:2105.04906 [cs.CV]. [Online]. Available: \url{http://arxiv.org/abs/2105.04906}
\BIBentrySTDinterwordspacing

\bibitem{assran_self-supervised_2023-1}
\BIBentryALTinterwordspacing
M.~Assran, Q.~Duval, I.~Misra, P.~Bojanowski, P.~Vincent, M.~Rabbat, Y.~LeCun, and N.~Ballas, ``Self-{Supervised} {Learning} from {Images} with a {Joint}-{Embedding} {Predictive} {Architecture},'' Apr. 2023, arXiv:2301.08243 [cs.CV]. [Online]. Available: \url{http://arxiv.org/abs/2301.08243}
\BIBentrySTDinterwordspacing

\bibitem{oord_representation_2019}
\BIBentryALTinterwordspacing
A.~van~den Oord, Y.~Li, and O.~Vinyals, ``Representation {Learning} with {Contrastive} {Predictive} {Coding},'' Jan. 2019, arXiv:1807.03748. [Online]. Available: \url{http://arxiv.org/abs/1807.03748}
\BIBentrySTDinterwordspacing

\bibitem{detone_superpoint_2018}
\BIBentryALTinterwordspacing
D.~DeTone, T.~Malisiewicz, and A.~Rabinovich, ``{SuperPoint}: {Self}-{Supervised} {Interest} {Point} {Detection} and {Description},'' Apr. 2018, arXiv:1712.07629 [cs.CV]. [Online]. Available: \url{http://arxiv.org/abs/1712.07629}
\BIBentrySTDinterwordspacing

\bibitem{sheeny_radiate_2021}
\BIBentryALTinterwordspacing
M.~Sheeny, E.~D. Pellegrin, S.~Mukherjee, A.~Ahrabian, S.~Wang, and A.~Wallace, ``{RADIATE}: {A} {Radar} {Dataset} for {Automotive} {Perception} in {Bad} {Weather},'' Apr. 2021, arXiv:2010.09076 [cs.CV]. [Online]. Available: \url{http://arxiv.org/abs/2010.09076}
\BIBentrySTDinterwordspacing

\bibitem{kim_mulran_2020}
\BIBentryALTinterwordspacing
G.~Kim, Y.~S. Park, Y.~Cho, J.~Jeong, and A.~Kim, ``{MulRan}: {Multimodal} {Range} {Dataset} for {Urban} {Place} {Recognition},'' in \emph{2020 {IEEE} {International} {Conference} on {Robotics} and {Automation} ({ICRA})}, May 2020, pp. 6246--6253, iSSN: 2577-087X. [Online]. Available: \url{https://ieeexplore.ieee.org/document/9197298}
\BIBentrySTDinterwordspacing

\bibitem{gadd_oord_2024}
\BIBentryALTinterwordspacing
M.~Gadd, D.~D. Martini, O.~Bartlett, P.~Murcutt, M.~Towlson, M.~Widojo, V.~Muşat, L.~Robinson, E.~Panagiotaki, G.~Pramatarov, M.~A. Kühn, L.~Marchegiani, P.~Newman, and L.~Kunze, ``{OORD}: {The} {Oxford} {Offroad} {Radar} {Dataset},'' May 2024, arXiv:2403.02845 [cs.RO]. [Online]. Available: \url{http://arxiv.org/abs/2403.02845}
\BIBentrySTDinterwordspacing

\bibitem{jiang_go_2026}
\BIBentryALTinterwordspacing
P.~Jiang, K.~Viswanath, A.~Nagariya, G.~Chustz, M.~Wigness, P.~Osteen, T.~Overbye, C.~Ellis, L.~Quang, J.~Huang, and S.~Saripalli, ``{GO}: {The} {Great} {Outdoors} {Multimodal} {Dataset},'' June 2026, arXiv:2501.19274 [cs.RO]. [Online]. Available: \url{http://arxiv.org/abs/2501.19274}
\BIBentrySTDinterwordspacing

\bibitem{barnes_oxford_2020}
\BIBentryALTinterwordspacing
D.~Barnes, M.~Gadd, P.~Murcutt, P.~Newman, and I.~Posner, ``The {Oxford} {Radar} {RobotCar} {Dataset}: {A} {Radar} {Extension} to the {Oxford} {RobotCar} {Dataset},'' Feb. 2020, arXiv:1909.01300 [cs.RO]. [Online]. Available: \url{http://arxiv.org/abs/1909.01300}
\BIBentrySTDinterwordspacing

\bibitem{kim_hercules_2025}
\BIBentryALTinterwordspacing
H.~Kim, M.~Jung, C.~Noh, S.~Jung, H.~Song, W.~Yang, H.~Jang, and A.~Kim, ``{HeRCULES}: {Heterogeneous} {Radar} {Dataset} in {Complex} {Urban} {Environment} for {Multi}-session {Radar} {SLAM},'' Feb. 2025, arXiv:2502.01946 [cs.RO]. [Online]. Available: \url{http://arxiv.org/abs/2502.01946}
\BIBentrySTDinterwordspacing

\bibitem{boxan_fomo_2026}
\BIBentryALTinterwordspacing
M.~Boxan, G.~Jeanson, A.~Krawciw, E.~Daum, X.~Qiao, S.~Lilge, T.~D. Barfoot, and F.~Pomerleau, ``{FoMo}: {A} {Multi}-{Season} {Dataset} for {Robot} {Navigation} in {Forêt} {Montmorency},'' Mar. 2026, arXiv:2603.08433 [cs.RO]. [Online]. Available: \url{http://arxiv.org/abs/2603.08433}
\BIBentrySTDinterwordspacing

\bibitem{li_get_2025}
\BIBentryALTinterwordspacing
S.~Li, Z.~Hong, Y.~Chen, L.~Hu, and J.~Qin, ``Get {It} for {Free}: {Radar} {Segmentation} {Without} {Expert} {Labels} and {Its} {Application} in {Odometry} and {Localization},'' \emph{IEEE Robotics and Automation Letters}, vol.~10, no.~3, pp. 2678--2685, Mar. 2025. [Online]. Available: \url{https://ieeexplore.ieee.org/document/10857300}
\BIBentrySTDinterwordspacing

\bibitem{vizzo_kiss-icp_2023}
\BIBentryALTinterwordspacing
I.~Vizzo, T.~Guadagnino, B.~Mersch, L.~Wiesmann, J.~Behley, and C.~Stachniss, ``{KISS}-{ICP}: {In} {Defense} of {Point}-to-{Point} {ICP} – {Simple}, {Accurate}, and {Robust} {Registration} {If} {Done} the {Right} {Way},'' \emph{IEEE Robotics and Automation Letters}, vol.~8, no.~2, pp. 1029--1036, Feb. 2023. [Online]. Available: \url{https://ieeexplore.ieee.org/document/10015694}
\BIBentrySTDinterwordspacing

\bibitem{korovko_cuvslam_2025}
\BIBentryALTinterwordspacing
A.~Korovko, D.~Slepichev, A.~Efitorov, A.~Dzhumamuratova, V.~Kuznetsov, H.~Rabeti, J.~Biswas, and S.~Pouya, ``\BIBforeignlanguage{en}{{cuVSLAM}: {CUDA} accelerated visual odometry and mapping},'' June 2025. [Online]. Available: \url{https://arxiv.org/abs/2506.04359v3}
\BIBentrySTDinterwordspacing

\bibitem{boxan_radar_2026}
\BIBentryALTinterwordspacing
M.~Boxan, W.~Larrivée-Hardy, and F.~Pomerleau, ``Radar {Odometry} {Subject} to {High} {Tilt} {Dynamics} of {Subarctic} {Environments},'' Apr. 2026, arXiv:2604.19962 [cs.RO]. [Online]. Available: \url{http://arxiv.org/abs/2604.19962}
\BIBentrySTDinterwordspacing

\end{thebibliography}
}
\end{document}